\documentclass[runningheads]{eccv_files/llncs}

\usepackage{eccv_files/eccv}

\usepackage{eccv_files/eccvabbrv}

\usepackage{graphicx}
\usepackage{booktabs}
\usepackage{bm}
\usepackage{colortbl}
\usepackage{multirow}
\usepackage{pifont}
\usepackage{amsfonts}

\usepackage[accsupp]{axessibility}  

\usepackage[pagebackref,breaklinks,colorlinks,citecolor=eccvblue]{hyperref}

\usepackage{orcidlink}

\begin{document}

\title{Learning Non-Linear Invariants for Unsupervised Out-of-Distribution Detection} 


\author{Lars Doorenbos\orcidlink{0000-0002-0231-9950} \and
Raphael Sznitman\orcidlink{0000-0001-6791-4753} \and
Pablo Márquez-Neila\orcidlink{0000-0001-5722-7618}}
\authorrunning{L. Doorenbos et al.}


\institute{
  University of Bern, Bern, Switzerland \\
  \texttt{\{lars.doorenbos,raphael.sznitman,pablo.marquez\}@unibe.ch}}

\maketitle

\begin{abstract} 
  The inability of deep learning models to handle data drawn from unseen distributions has sparked much interest in unsupervised out-of-distribution (U-OOD) detection, as it is crucial for reliable deep learning models. Despite considerable attention, theoretically-motivated approaches are few and far between, with most methods building on top of some form of heuristic. Recently, U-OOD was formalized in the context of data invariants, allowing a clearer understanding of how to characterize U-OOD, and methods leveraging affine invariants have attained state-of-the-art results on large-scale benchmarks. Nevertheless, the restriction to affine invariants hinders the expressiveness of the approach. In this work, we broaden the affine invariants formulation to a more general case and propose a framework consisting of a normalizing flow-like architecture capable of learning non-linear invariants. Our novel approach achieves state-of-the-art results on an extensive U-OOD benchmark, and we demonstrate its further applicability to tabular data. Finally, we show our method has the same desirable properties as those based on affine invariants.
\end{abstract}


\newif\ifdraft
\drafttrue

\definecolor{orange}{rgb}{1,0.5,0}
\definecolor{gr}{rgb}{0,0.65,0}
\definecolor{mygray}{gray}{0.95}

\ifdraft
 \newcommand{\RS}[1]{{\color{red}{\bf RS: #1}}}
 \newcommand{\rs}[1]{{\color{red}#1}}
 \newcommand{\PMN}[1]{{\color{orange}{\bf PMN: #1}}}
 \newcommand{\pmn}[1]{{\color{orange}#1}}
 \newcommand{\LD}[1]{{\color{blue}{\bf LD: #1}}}
 \newcommand{\ld}[1]{{\color{blue}#1}}
 \newcommand{\old}[1]{{\color{gr}#1}}
\else
 \renewcommand{\sout}[1]{}
 \newcommand{\RS}[1]{{\color{red}{}}}
 \newcommand{\rs}[1]{#1}
 \newcommand{\PMN}[1]{{\color{red}{}}}
 \newcommand{\pmn}[1]{#1}
\fi

\newcommand{\real}{\mathbb{R}}
\newcommand{\x}{\mathbf{x}}
\newcommand{\z}{\mathbf{z}}
\newcommand{\y}{\mathbf{y}}
\newcommand{\haty}{\hat{\y}}
\newcommand{\w}{\mathbf{w}}
\renewcommand{\d}{\mathbf{d}}
\newcommand{\D}{\mathcal{D}}
\newcommand{\X}{\mathcal{X}}
\newcommand{\Z}{\mathcal{Z}}
\newcommand{\J}{\mathbf{J}}
\newcommand{\bZ}{\mathbf{Z}}
\newcommand{\M}{\mathcal{M}}
\newcommand{\I}{\mathcal{I}}
\newcommand{\jacobian}{\mathbf{J}}
\newcommand{\balpha}{\bm{\alpha}}
\newcommand{\pkde}{p_{\textrm{kde}}}
\newcommand{\psv}{p_{\balpha}}
\newcommand{\f}{\mathbf{f}}
\newcommand{\g}{\mathbf{g}}
\newcommand{\F}{\mathcal{F}}
\renewcommand{\a}{\mathbf{a}}

\newcommand{\MSCL}{{\bf{MSCL}}}
\newcommand{\PANDA}{{\bf{PANDA}}}
\newcommand{\MKD}{{\bf{MKD}}}
\newcommand{\SSD}{{\bf{SSD}}}
\newcommand{\DNtwo}{{\bf{DN2}}}
\newcommand{\MHRot}{{\bf{MHRot}}}
\newcommand{\DDV}{{\bf{DDV}}}
\newcommand{\IC}{{\bf{IC}}}
\newcommand{\HierAD}{{\bf{HierAD}}}
\newcommand{\Glow}{{\bf{Glow}}}
\newcommand{\MahaAD}{{\bf{MahaAD}}}
\newcommand{\CFlow}{{\bf{CFlow}}}
\newcommand{\NL}{{\bf{NL-Invs}}}
\newcommand{\DIF}{{\bf{DIF}}}

\newcommand{\uniclass}{\emph{uni-class}}
\newcommand{\uniano}{\emph{uni-ano}}
\newcommand{\unimed}{\emph{uni-med}}
\newcommand{\shiftlowres}{\emph{shift-low-res}}
\newcommand{\shifthighres}{\emph{shift-high-res}}

\newcommand{\thyroid}{\emph{thyroid}}
\newcommand{\bc}{\emph{breast cancer}}
\newcommand{\speech}{\emph{speech}}
\newcommand{\pg}{\emph{pen global}}
\newcommand{\shuttle}{\emph{shuttle}}
\newcommand{\kdd}{\emph{KDD99}}

\newcommand{\guood}{{\bf{General U-OOD}}}
\newcommand{\shd}{{\bf{Shallow U-OOD}}}

\newcommand{\xmark}{\ding{55}}%
\newcommand{\cmark}{\ding{51}}%

\section{Introduction}

Deep learning (DL) models can perform remarkably in controlled settings, where samples evaluated come from the same distribution as those seen during training. Unsurprisingly, real-world scenarios rarely allow for such controlled settings, and a mismatch between train and test distributions is often a reality instead. Additionally, evaluating \textit{out-of-distribution} (OOD) samples comes with few guarantees, and model performance is typically poorer than expected. More insidiously, no obvious in-built way exists to identify when the evaluated sample differs from the training distribution. Jointly, these shortcomings limit the use of DL models in real-world settings, as their reliability cannot be taken for granted.

Consequently, OOD samples need to be detected beforehand to ensure that unreliable model predictions for those samples can be dealt with appropriately. This problem has become known as OOD detection~\cite{hendrycks2016baseline} and shares goals with related fields such as anomaly detection, novelty detection, outlier detection, one-class classification, and open-set recognition~\cite{salehi2021unified}. Here, we consider generalized OOD~\cite{yang2021generalized}, where any distributional shift from the in-distribution should be identified.

\begin{figure}[t]
    \centering
    \includegraphics[width=0.9\linewidth]{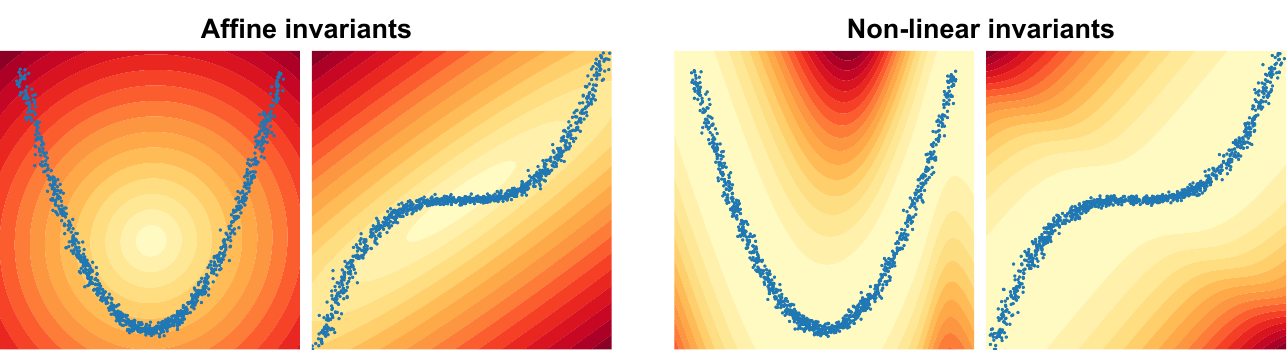}
    \caption{\textbf{Motivation for learning non-linear invariants.} Affine functions (left) are not expressive enough to model the invariants of the data and are thus unsuccessful at OOD detection. Instead, non-linear functions~(right) are more general and flexible. Blue points indicate training samples; darker colors denote regions with higher OOD scores.}
    \label{fig:intro}
\end{figure}

OOD detection can be divided into supervised and unsupervised OOD (U-OOD). Supervised OOD methods can access the labels of a downstream task or explicit OOD samples. In contrast, U-OOD methods operate solely on unlabeled training samples. The lack of training labels or OOD samples is an important reason why U-OOD is so challenging, as determining what should be considered OOD is not always clear. Unlike the supervised case, one cannot rely on marking every sample that does not belong to one of the classes as OOD. To address this,~\cite{doorenbos2022data} proposed characterizing datasets with multiple data invariants. Specifically, data points that do not have the expected value for any of these invariants are deemed OOD. With this characterization, it is possible to assess what datasets can be used to evaluate U-OOD detectors by considering whether a potential dataset satisfies all the invariants in the training data. Formally, the data invariants characterization of U-OOD aims to define a set of functions over the training features with a \mbox{(near-)constant} value. The union of these functions is used at inference time to spot U-OOD samples by testing whether the invariants hold for a given new sample. When restricting invariants to affine functions, the problem can be cast in terms of principal component analysis (PCA) and achieves state-of-the-art results on a large-scale benchmark~\cite{doorenbos2022data}. 

However, it seems improbable that affine functions are sufficient to characterize all invariants present in training datasets. Examples of their limitations are easily found, as exemplified in Fig.~\ref{fig:intro}. Despite the potential benefits of non-linear invariants for U-OOD detection, their actual advantages are still unexplored. In this work, we propose to find non-linear invariants by modeling them with a \emph{volume preserving network}, a bijective function inspired by normalizing flows that deforms the input space while preserving the volume almost everywhere by design. Since the network cannot perform a projection, any invariant dimension at the network's output when processing the training data must necessarily be an invariant of the training data. We extensively evaluate our approach and demonstrate that non-linear invariants outperform previous U-OOD detection methods. Moreover, we show how our method extends to different modalities by its application to tabular data and its benefit over affine invariants. 

In summary, our main contributions are (1)~a generalization of the invariant-based characterization of U-OOD that allows for the inclusion of non-linearities, (2) a novel embodiment of this framework that can learn non-linear invariants, and (3)~an extensive evaluation of our method and other state-of-the-art methods on two benchmarks, the large image benchmark from~\cite{doorenbos2022data} and a novel tabular benchmark.

\section{Related work}

While developing new supervised OOD detection methods is an active area of research (\emph{e.g.} \cite{du2022vos,hendrycks2016baseline,hsu2020generalized,katz2022training,ming2022exploit,lee2018simple,liang2017enhancing,liu2020energy,sun2022out}), their reliance on labeled datasets and trained classifiers limit their applicability. For the remainder of this section, we focus on unsupervised approaches.

Generative models have played an important role in U-OOD. In theory, generative models make for excellent U-OOD detectors because of their capability to estimate complex data distributions. However, in practice, they fail even in straightforward cases \cite{choi2018waic,marquez2019image,nalisnick2018deep,serra2019input}. Various explanations and remedies for this have been proposed, based on, for instance, input complexity \cite{serra2019input}, background information \cite{ren2019likelihood,xiao2020likelihood}, architectural limitations \cite{kirichenko2020normalizing}, ensembles \cite{choi2018waic}, or typicality \cite{morningstar2021density,nalisnick2019detecting,osada2023out}. Most recently, approaches based on diffusion models have gained popularity~\cite{liu2023unsupervised,pinaya2022fast,shi2023dissolving,wyatt2022anoddpm}, although they also require heuristics to function, as using the estimated data likelihood is often insufficient.

Alternatively, representation learning-based methods have been proposed for U-OOD. Here, a model is trained using a self-supervised approach, and a test sample is scored using the model's output probabilities~\cite{bergman2020classification,hendrycks2019using}, or by a simple anomaly detector operating on the features of the model~\cite{chen2023cluster,sehwag2021ssd,tack2020csi}. Initially, the self-supervised training task consisted of transformation prediction \cite{bergman2020classification,hendrycks2019using}, while more recent methods use contrastive learning \cite{chen2023cluster,sehwag2021ssd,tack2020csi}. 

Rather than training a model with a self-supervised training task, state-of-the-art methods use a network pre-trained on a general dataset, such as ImageNet, to provide a strong foundation for the U-OOD task. Using these features directly already provides high performance \cite{bergman2020deep,luan2021out,ouardini2019towards,rippel2021modeling}, while other works adapt these features to a target domain using an OOD-specific loss function \cite{marquez2019image,reiss2021panda,reiss2021mean}. Our work follows the first line of work, relying on the features of a frozen pre-trained model. We use ResNet architectures to run competing baselines and facilitate comparisons with earlier works. However, our method is in no way restricted to this architectural choice.

Architecturally, our approach is closely linked to normalizing flows (NF)~\cite{kingma2018glow}. More precisely, our method resembles an NF where we only have volume-preserving operations and lack the generative objective. These choices set us apart from other NF-based OOD works~\cite{chali2023improving,kirichenko2020normalizing,schirrmeister2020understanding,serra2019input} and are validated by our experiments. A closely related method from this field is the Denoising Normalizing Flow (DNF) model~\cite{horvat2021denoising}. Proposed for an entirely different purpose, the DNF aims to find a low-dimensional manifold dataset embedding and estimate the density of samples in this low-dimensional space. The DNF is trained with the standard generative objective alongside a reconstruction error term. After the forward pass, a predetermined number of output dimensions are set to~0 before reversing through the network. While the DNF ignores these dimensions, our approach forces them to be invariant and uses them as a scoring method for OOD samples.
Furthermore, the DNF is not volume-preserving. Some works do exist on volume-preserving neural networks \cite{macdonald2021volume,zhu2022vpnets}, but these are designed for entirely different purposes, such as classification, and are thus very different in design. 


\section{Method}
\label{sec:method}

Given a training set~$\{\x_i\}_{i=1}^N$, with corresponding feature vectors~$\f(\x_i) \equiv \f_i \in \real^D$, we define an invariant following~\cite{doorenbos2022data} as a non-constant function,~$g:\real^D\to\real$, such that $g(\f_i)=0,\ \forall i$. That is, $g$~is an invariant if it computes a constant value (\ie,~$g(\f_i)=0$) for the training set elements but may compute different constant values for other elements. For convenience, we will stack the invariants in a single vector function~$\g:\real^D\to\real^K$ with~$\g=(g_1, \ldots, g_K)$. Our goal is to find a function~$\g$ of invariants that satisfies
\begin{align}
    \label{eq:invariant_g}
    \g(\f_i)=\mathbf{0} &\quad \forall i, \\
    \label{eq:independent_g}
    \det(\J(\f_i)\cdot\J^T(\f_i)) \neq 0 &\quad \forall i,
\end{align}
where $\J(\f_i)$~is the Jacobian of~$\g$ evaluated at~$\f_i$. The second condition ensures that no component of~$\g$ is trivially constant and that there are no redundant invariants by making the Jacobian~$\J$ full rank. The 0~level-set of~$\g$ that satisfies these conditions defines an implicit manifold on the feature space~$\real^D$. A test feature vector~$\f$ will be considered OOD if it does not lie on the manifold (\ie,~$\g(\f) \neq \mathbf{0}$).

However, noisy real-world data rarely lies on an exact manifold, and solving Eq.~\eqref{eq:invariant_g} for a reasonably regularized~$\g$ is unfeasible even for a small number of invariants~$K$ in practice. Instead, as proposed in~\cite{doorenbos2022data}, we relax these conditions and find a set of \emph{soft invariants} (\ie,~functions that are approximately constant for all the training set elements). These are found by optimizing a soft version of Eq.~\eqref{eq:invariant_g},
\begin{align}
    \label{eq:soft_invariant_g}
    \min_\g & \sum_i \|\g(\f_i)\|^2_2 \\
    \label{eq:soft_independent_g}
    \textrm{s.t.} & \det(\J(\f_i)\cdot\J^T(\f_i)) \neq 0 \quad \forall i.
\end{align}

Once the function~$\g$ is found, test feature vectors are evaluated by measuring how much they violate each invariant compared to the elements of the training set. Specifically, a test vector~$\f$ is scored by computing the ratios between the test squared error and the average training squared error,
\begin{equation}
    \label{eq:score_g}
    s(\f) = \sum_{k=1}^K\dfrac{g_k(\f)^2}{e_k},
\end{equation}
where $e_k$ is the mean squared error of the soft invariant~$g_k$ on the training set,
\begin{equation}
    \label{eq:training_error}
    e_k = \dfrac{1}{N}\sum_i g_k(\f_i)^2.
\end{equation}
Intuitively, strong invariants with low~$e_k$ values will strongly influence the final score, while weak invariants with large~$e_k$ values will effectively be ignored. 

\begin{figure}[t]
    \centering
    \includegraphics[width=0.5\linewidth]{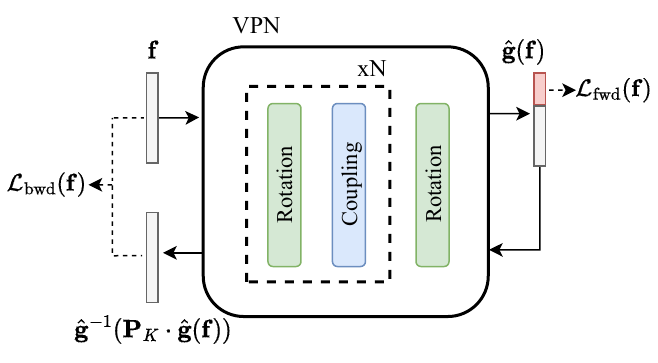}
    \caption{\textbf{Architecture of our proposed volume preserving network.} The VPN is a fully invertible model with alternating rotation and coupling layers.}
    \label{fig:method}
\end{figure}

The work~\cite{doorenbos2022data} simplified the problem by modelling invariants as affine functions~$\g(\f)=\mathbf{A}\f+\mathbf{b}$, which allowed for tractable solutions of Eq.~\eqref{eq:soft_invariant_g}. Specifically, it was shown that finding $\mathbf{A}$ and $\mathbf{b}$ could be done by applying PCA to the training features and that Eq.~\eqref{eq:score_g} was equivalent to the square of the Mahalanobis distance.

\begin{figure*}[t]
    \centering
    \setlength\tabcolsep{8pt}
    \begin{tabular}{ccc}
        \includegraphics[width=0.3\linewidth]{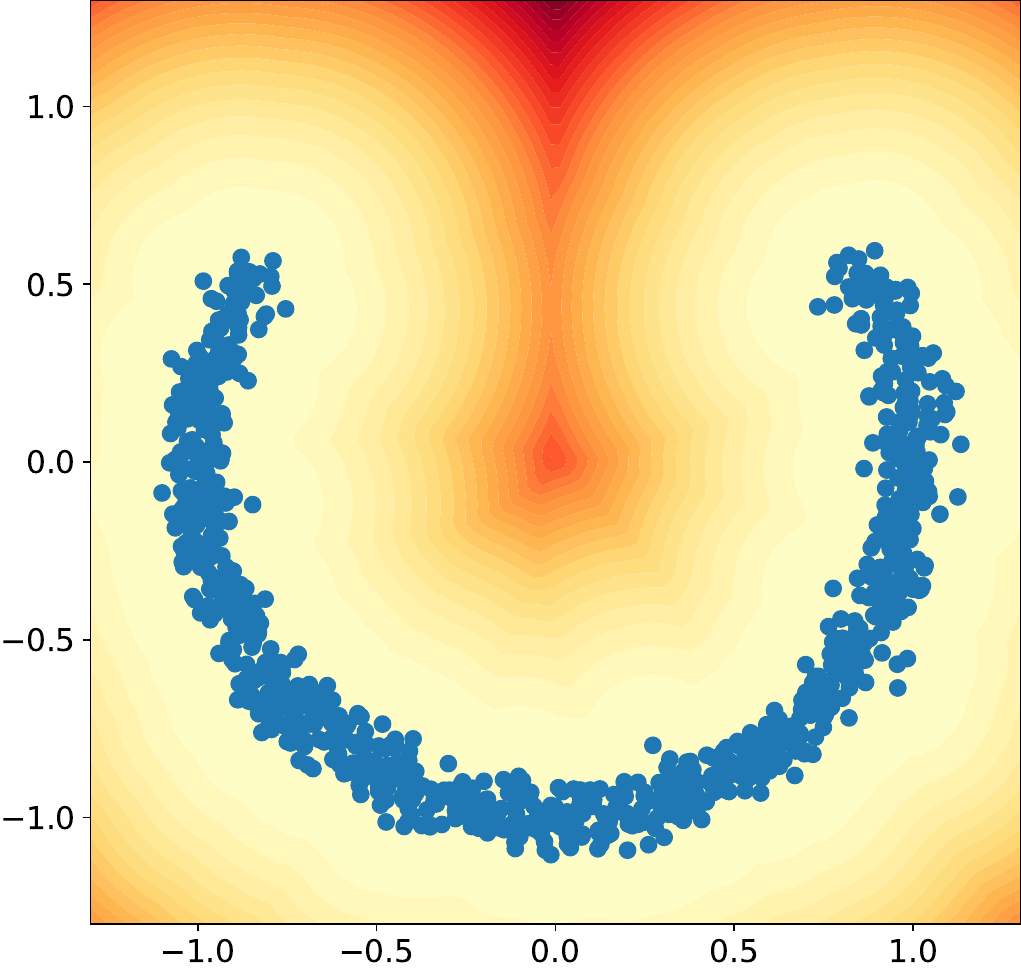} &
        \includegraphics[width=0.3\linewidth]{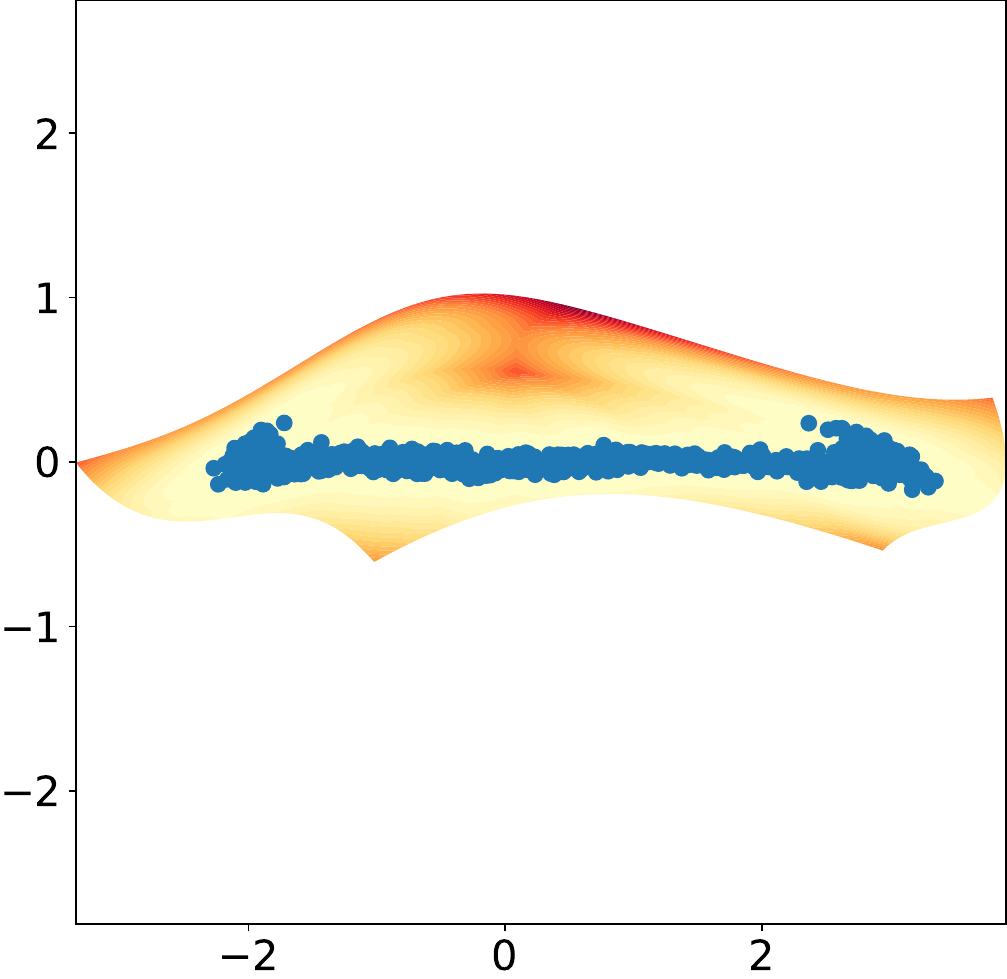} &
        \includegraphics[width=0.3\linewidth]{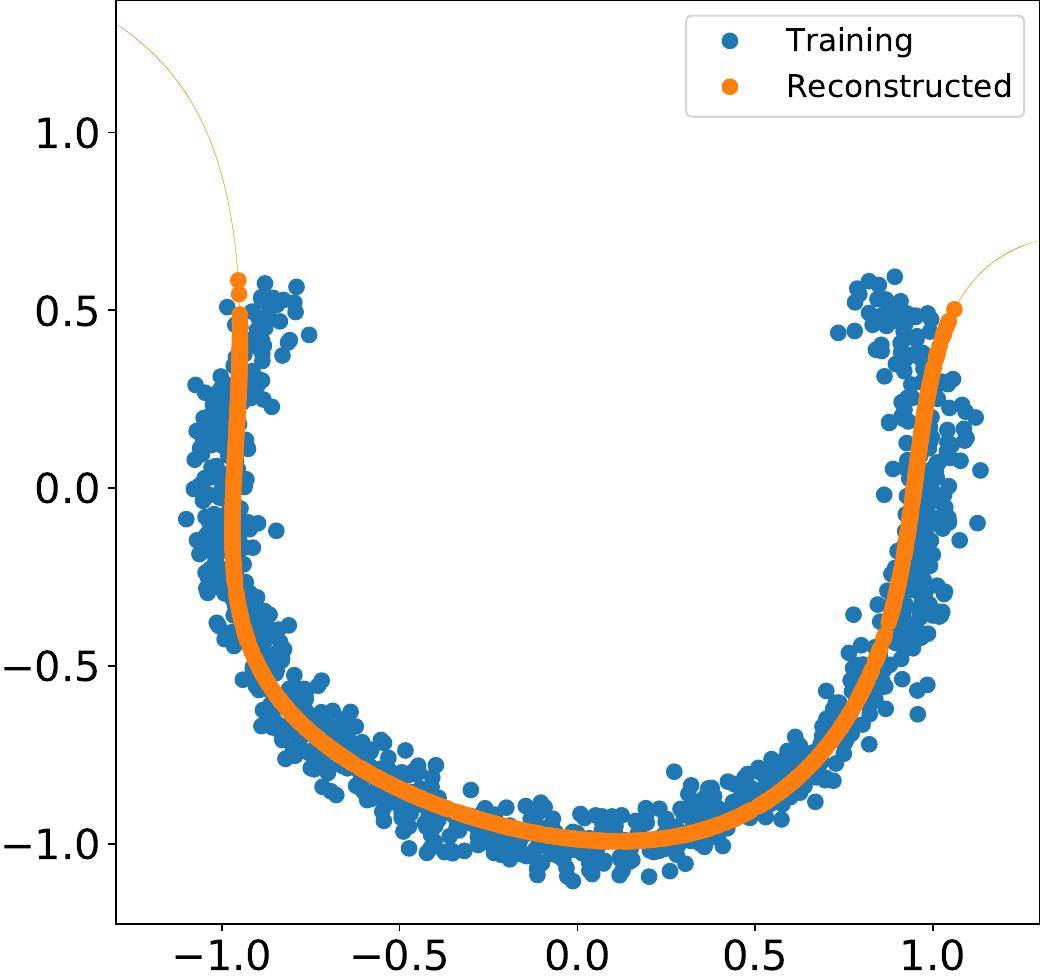} \\
        (a) & (b) & (c) \\
        \multicolumn{3}{c}{\includegraphics[width=\linewidth]{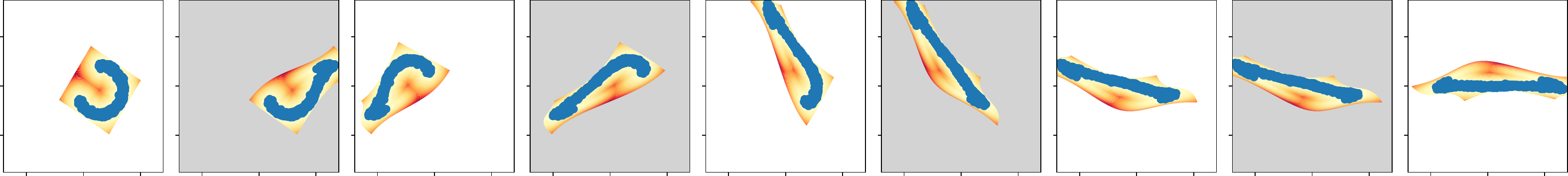}}
    \end{tabular} 
    \caption{\textbf{Example of finding non-linear invariants with the VPN on a toy dataset.} (a)~illustrates the data, (b)~the invariant representation, and (c)~the reconstruction of the training data from the invariant representation after zeroing the invariant dimension together with the original data. Background color indicates the distance to the nearest training data point in the original space and tracks how these are modified after the forward and backward pass. In~(c), this is compressed into a thin, barely visible line from both ends of the U~shape. The images below show how the data is transformed through the nine layers of the network. Images with a white-shaded background result from rotation layers, and images with a gray background result from coupling layers.}
    \label{fig:toy_data}
\end{figure*}

\subsection{Non-linear invariants}

In this work, we relax the assumption of affine invariants and allow for a broader family of invariants by modeling the function~$\g$ with a deep neural network~$\hat{\g}$.
Specifically, we impose the constraint of Eq.~\eqref{eq:soft_independent_g} in the neural network design by choosing an architecture that ensures full-rank Jacobians. 
Inspired by normalizing flows~\cite{kingma2018glow}, we design a \emph{volume preserving network}~(VPN) as a bijective function~$\hat{\g}:\real^D\to\real^D$ composed of bijective operations with unimodular Jacobians. A volume-preserving approach prevents the network from learning a projection to a (near-)constant value, which would artificially create invariants. Preserving the volume forces the network to learn actual invariants instead of shortcuts.

In particular, we design our VPN by alternating rotation and coupling layers. Rotation layers are linear layers with orthogonal transformations and a bias vector. We parameterize an orthogonal layer of $n$~dimensions with a $\binom{n}{2}$-dimensional vector~$\mathbf{v}$ and an $n$-dimensional bias vector~$\mathbf{b}$. The layer transforms an input vector~$\x$ as,
\begin{equation}
    r(\x) = e^{[\mathbf{v}]_\times}\cdot \x + \mathbf{b},
\end{equation}
where $[\mathbf{v}]_\times$~is the skew symmetric matrix with the elements of~$\mathbf{v}$, and $e$~is the matrix exponential. The Jacobian of an orthogonal layer is the orthogonal matrix~$e^{[\mathbf{v}]_\times}$ and has, therefore, determinant~$1$. Coupling layers~\cite{dinh2014nice} use some of the components of the input vector to compute a transformation that will be applied to the remaining components,
\begin{align}
    (\x_a, \x_b) &= \textrm{split}(\x), \\
    \nonumber
    \y &= \textrm{join}(\x_a + t(\x_b), \x_b),
\end{align}
\noindent
where~$\x$ and~$\y$ are the input and output of the coupling layer, respectively, and $t$~is a multi-layer perceptron~(MLP) computing a translation. Unlike~\cite{dinh2014nice,dinh2016density}, no scale factor is applied to keep the Jacobian unimodular. Both orthogonal and coupling layers are easily inverted. In particular, the inverse of an orthogonal layer is,
\begin{equation}
    r^{-1}(\y) = e^{[-\mathbf{v}]_\times}\cdot (\y - \mathbf{b}),
\end{equation}
and for the coupling layer,
\begin{align}
    (\y_a, \y_b) &= \textrm{split}(\y), \\
    \x &= \textrm{join}(\y_a - t(\y_b), \y_b). \nonumber
\end{align}

The composition of alternating rotation and coupling layers ensures that the complete VPN~$\hat{\g}$ is an invertible function with unimodular Jacobian and is, therefore, volume-preserving almost everywhere. The invariant function~$\g:\real^D\to\real^K$ is defined by the first $K$ outputs of the VPN,~$\g=\hat{\g}_{1:K}$. Its Jacobian~$\J$, corresponding to the first $K$~rows of the Jacobian of~$\hat{\g}$, is also full rank, thus satisfying the constraint of Eq.~\eqref{eq:soft_independent_g} by design. Eq.~\eqref{eq:soft_invariant_g} can now be solved efficiently by simply minimizing the \emph{forward loss},
\begin{equation}
    \mathcal{L}_{\textrm{fwd}}(\f) = \|\hat{\g}_{1:K}(\f)\|_2^2.
\end{equation}



In addition, we leverage the bijectivity of~$\hat{\g}$ to define a \emph{backward loss} minimizing the reconstruction error between a training feature vector~$\f$ and its reconstruction,
\begin{equation}
    \label{eq:loss_bwd}
    \mathcal{L}_{\textrm{bwd}}(\f) = \|\hat{\g}^{-1}\left(\mathbf{P}_K\cdot\hat{\g}(\f)\right) - \f\|_2^2,
\end{equation}
where $\mathbf{P}_K$~is a diagonal linear operator projecting the first $K$~dimensions to~$0$, which zeroes the invariants. Although optimizing the forward loss implicitly minimizes the backward loss, we found that explicitly introducing the backward loss improved the stability of the training and the performance in our experiments. Nonetheless, the backward loss by itself also encodes invariants: by reconstructing the data from a representation where $K$ dimensions are zeroed out with a volume-preserving network, all variance must be in the non-invariant dimensions for a good reconstruction, and the $K$ zeroed dimensions will encode invariants. The final training loss is the sum of the forward and backward losses. A schematic of our approach can be found in Fig.~\ref{fig:method}

To illustrate our approach, we use the 2-dimensional toy example depicted in Figure~\ref{fig:toy_data}. The data shown in Figure~\ref{fig:toy_data}(a) has no affine invariant (\ie,~there exists no affine~$g_k$ for which $\dfrac{1}{N}\sum_i g_k(\x_i)^2$ is close to~0). However, it does have a soft non-linear invariant, namely, the distance of the samples to the origin. We therefore set~$K=1$.

After training, we pass the data through the network to obtain an invariant representation shown in Figure~\ref{fig:toy_data}(b). The network has learned an almost constant dimension for the training data, the non-linear invariant, and the variability is encoded in the other dimension. On the other hand, the OOD~samples are not invariant along this dimension and score higher than in-distribution samples when compared with Eq.~\eqref{eq:score_g}. 

Figure~\ref{fig:toy_data}(c) shows the result of reconstructing the data with the composition $\hat{\g}^{-1}\circ\mathbf{P}_K\circ\hat{\g}$ from Eq.~\eqref{eq:loss_bwd}. After zeroing the invariant with~$\mathbf{P}_K$, the reconstructed data lies in a one-dimensional manifold that minimizes the distance to the original data and reduces the backward loss while removing noise in the radial direction. Therefore, the invariant measures deviations from this manifold.

\subsection{Multi-scale invariants}
\label{sec:multi-scale}

As in~\cite{doorenbos2022data}, we use a pre-trained CNN to compute feature descriptors at multiple scales. The CNN is applied to each input image~$\x$ to generate a collection of feature vectors~$\{\f_\ell(\x)\}_{\ell=1}^L$ by performing global average pooling on the activation maps at each layer~$\ell$. During training, the training feature vectors $\{\f_\ell(\x_i)\}_{i=1}^N$ at layer~$\ell$ are used to train a set of $L$~invariant functions~$\{\g^{(\ell)}\}_{\ell=1}^L$ through the procedure described in the previous section. Each function~$\g^{(\ell)}$ is trained with a different number of invariants~$K_\ell$, which are hyperparameters of our method.

At inference time, the test images~$\x$ are evaluated by computing layer-wise scores~$s_\ell(\f_\ell(\x))$ following Eq.~\eqref{eq:score_g},
\begin{equation}
    s_\ell(\f) = \sum_{k=1}^{K_\ell} \dfrac{g_k^{(\ell)}(\f)}{e_k^{(\ell)}},
\end{equation}
which are aggregated to compute the final invariant score,
\begin{equation}
\label{eq:inv_score}
    S_{\text{inv}}(\x) = \sum_{\ell=1}^L s_\ell(\f_\ell(\x)).
\end{equation}

\subsection{Scoring samples}
We empirically found our invariant score of Eq.~\eqref{eq:inv_score} to be complementary to a standard 2-NN score~\cite{bergman2020deep} and observed that combining the two scores leads to a further boost in performance. To compute the 2-NN score, we first define the 2-NN distance of a test sample at a layer~$\ell$ as 
\begin{equation}
    \text{dist-2nn}_\ell(\f) = \frac{1}{2}\sum_{\f_n\in N_2^{(\ell)}(\f)}\|\f-\f_n\|_2,
\end{equation}
where $N_2^{(\ell)}(\f)$ are the $2$ nearest neighbours of $\f$ in the training set at layer $\ell$.
As with the layer-wise invariant score, the 2-NN~distances are normalized by the average 2-NN~distances of the training set,
\begin{equation}
    \textrm{s-2nn}_\ell(\f) = K_\ell \frac{\text{dist-2nn}(\f)}{\frac{1}{N}\sum_i \text{dist-2nn}(\f_\ell(\x_i))},
\end{equation}
where the factor~$K_\ell$ compensates for the difference in magnitude with respect to the invariant score~$s_\ell$. In the denominator, the 2-NN~distances are calculated for the training set elements to themselves, making each feature vector~$\f_\ell(\x_i)$ its own first neighbor. To avoid this, we exclude the element~$\f_\ell(\x_i)$ from the training set when computing~$\text{dist-2nn}(\f_\ell(\x_i))$. The 2NN~score is computed as,
\begin{equation}
    S_{\text{2nn}}(\x) = \sum_{\ell=1}^L \textrm{s-2nn}_\ell(\f_\ell(\x)),
\end{equation}
and the final score is the sum of the invariant and the 2NN~scores,
\begin{equation}
    S_{\text{final}}(\x) = S_{\text{inv}}(\x) + S_{\text{2nn}}(\x).
\end{equation}
We will analyze the contribution of each of these terms to the detection performance in the ablation study of the results section.

\section{Experiments}

\subsection{Benchmarks}
We use the U-OOD evaluation benchmark introduced in~\cite{doorenbos2022data} and propose a new benchmark with shallow datasets for additional experiments. Both benchmarks are described below.

\textbf{\guood.} The U-OOD benchmark introduced in~\cite{doorenbos2022data} consists of 73 experiments spread over five tasks, each containing varying criteria for the in and out distributions. Three of the tasks have an unimodal training dataset: \uniclass, containing 30 one-class classification experiments on the low-resolution CIFAR10 and CIFAR100 datasets; \uniano, which consists of 15~experiments on the high-resolution MVTec images where the number of training images is limited; and \unimed, which has 7~experiments on different medical imaging modalities. The remaining two tasks use entirely different datasets as OOD. These are \shiftlowres, containing the CIFAR10:SVHN experiment on which many OOD-detectors fail, and \shifthighres, comprising 20~experiments with the DomainNet dataset. 
    
\textbf{\shd.} Collection of experiments on \emph{shallow} anomaly detection datasets with tabular data where deep neural network features from images are unavailable. This benchmark aims to show the generality of our approach to other data modalities. We use six tabular datasets from~\cite{goldstein2016comparative}. These datasets were conceived for unsupervised anomaly detection and contain inliers and outliers intertwined within the data. To adapt the datasets to our OOD detection problem, we pre-processed them by separating all the outliers and an equal number of inliers from each dataset and reserving them for the testing split. The remaining inliers were utilized as training data. The datasets included in the benchmark are \thyroid, \bc, \speech, \pg, \shuttle~and \kdd. Further details are provided in the appendix.

\subsection{Baselines}

For the \guood{} benchmark, we compare our method \NL{} against nine state-of-the-art methods. Six methods, \DNtwo~\cite{bergman2020deep}, \CFlow~\cite{gudovskiy2022cflow}, \DDV~\cite{marquez2019image}, \DIF~\cite{ouardini2019towards}, \MSCL~\cite{reiss2021mean}, and \MahaAD~\cite{rippel2021modeling} that use the same ResNet-101 backbone initialized with ImageNet pre-trained features, and three normalizing flow methods, \Glow~\cite{kingma2018glow}, \IC~\cite{serra2019input}, and \HierAD~\cite{schirrmeister2020understanding}.

For the \shd{} benchmark, we compare \NL{} to the baselines \MahaAD~(Mahalanobis distance), \DNtwo~(kNN), and \DIF~(Isolation Forest). The remaining baselines are bound to deep learning methods that cannot work with non-image or tabular data and are thus excluded from the comparison.

\subsection{Implementation details}

Our VPN architecture includes four rotation and coupling layers before the final rotation layer ($\text{N}=4$ in Fig.~\ref{fig:method}). Each coupling layer comprises an MLP with four linear layers of equal size as its input, interspersed with ReLU activations.

\NL{} requires setting the number of invariants per layer~$K_\ell$, as described in Sect.~\ref{sec:multi-scale}. Considering these values as independent hyperparameters would exponentially increase the search space and evaluation time. Instead, we set each~$K_\ell$ to the largest number of principal components of the data at layer~$\ell$ that jointly explain less than $p$\%~of the variance, where $p$~is a hyperparameter shared by all layers.

We utilized a ResNet-101 for the multi-scale feature extraction of Sect~\ref{sec:multi-scale}. We extract features from $L=3$ feature maps at the end of the last ResNet blocks. Following~\cite{reiss2021mean}, we normalize the feature vectors of the final layer to the unit norm for improved performance. In all our experiments, we train for 25 epochs with $p$ set to 5 and a batch size of $64$. We use the Adam optimizer~\cite{kingma2014adam} with a learning rate of $10^{-3}$ linearly decaying to $10^{-4}$ over the epochs.

\begin{table*}[t]
\caption{\textbf{Comparative evaluation on \guood.} We report the mean and standard deviation of the AUC over three runs. Baselines taken from~\cite{doorenbos2022data}. \textbf{Bold} and \underline{underlined} indicate the best and second best per column, respectively. On aggregate across the experiments, \NL~obtains the best performance.}
\label{tab:best101}
\begin{center}
\begin{tabular}{>{\bfseries}lcccccc}
\toprule
\textbf{Method}     & \textbf{\uniclass}& \textbf{\uniano}      & \textbf{\unimed}        & \textbf{\shiftlowres}   & \textbf{\shifthighres}         &\textbf{Mean}   \\
\midrule
Glow & 53.8\tiny{$\pm$0.1} &82.0\tiny{$\pm$2.5}  &55.8\tiny{$\pm$0.8} &8.8 & 34.5\tiny{$\pm$0.1}  & 47.0 \\
DDV   & 65.8\tiny{$\pm$1.4}  &65.5\tiny{$\pm$0.2}&60.3\tiny{$\pm$3.2}& 47.9\tiny{$\pm$6.6}& 63.9\tiny{$\pm$4.9} & 60.7\\
CFlow  &75.0\tiny{$\pm$0.0}&\textbf{95.7\tiny{$\pm$0.1}} &68.8\tiny{$\pm$0.3} &6.6\tiny{$\pm$0.2} & 61.8\tiny{$\pm$0.3} & 61.6\\
IC & 55.7\tiny{$\pm$0.1}  & 73.6\tiny{$\pm$2.6} &65.1\tiny{$\pm$0.5} & \underline{95.0} & 65.8\tiny{$\pm$0.1} & 71.0 \\
HierAD & 63.0\tiny{$\pm$0.4} &81.6\tiny{$\pm$2.1}  &72.5\tiny{$\pm$0.6} & 93.9 & 75.0\tiny{$\pm$0.3} & 77.2 \\
DN2  &91.2&86.2&\underline{76.7}&57.4& 76.0 &  77.5\\
DIF & 85.8\tiny{$\pm$0.3}& 81.8\tiny{$\pm$0.8} &72.1\tiny{$\pm$0.2} & 80.3\tiny{$\pm$4.5} & \underline{80.4}\tiny{$\pm$0.8}  &80.1 \\
MSCL  & \textbf{96.3\tiny{$\pm$0.0}} & 86.4\tiny{$\pm$0.0} & 75.2\tiny{$\pm$0.1} & 88.3\tiny{$\pm$0.0} &74.4\tiny{$\pm$0.0} & 84.1\\
MahaAD  & 92.4 & \underline{91.3} & 75.7 & 94.3 & 78.6 & \underline{86.5} \\
\rowcolor{mygray} NL-Invs & \underline{93.3}\tiny{$\pm$0.0} & 85.8\tiny{$\pm$0.0} & \textbf{77.2\tiny{$\pm$0.0}} & \textbf{97.8}\tiny{$\pm$0.1} & \textbf{85.5}\tiny{$\pm$0.1} & \textbf{87.9}\\
\bottomrule
\end{tabular} 
\end{center}
\end{table*}

\section{Results}

This section describes the results obtained on the two benchmarks, followed by a multi-faceted analysis of the behavior of our method.

\textbf{\guood.}
The performances of \NL~and the other methods are shown in Tab.~\ref{tab:best101}. Most methods behave inconsistently across the benchmark, with different methods scoring high for each task. For instance, \CFlow~is the best scoring method on \uniano~by a large margin. However, its high performance does not translate to the other experiments, where it is consistently among the lowest-scoring methods. \DNtwo~struggles on \shiftlowres but scores consistently well on the other tasks. \DIF~achieves decent performance overall but reaches the second-best score on \shifthighres. \MSCL~and \MahaAD~are generally good, with \MahaAD~being superior to \MSCL~on all cases except \uniano. However, \NL{} is consistently among the best-performing methods, reaching the highest mean score of the benchmark and the best score on \unimed, \shiftlowres~and \shifthighres. Moreover, it outperforms the normalizing flow methods \CFlow, \HierAD~and \IC~by large margins.

\begin{table*}[t]
\caption{\textbf{Comparative evaluation on \shd.} We report the mean and standard deviation of the AUC over five runs. Methods without a reported standard deviation are deterministic. \textbf{Bold} and \underline{underlined} indicate best and second best per column, respectively. \NL~performs best overall.}
\label{tab:shallow}
\begin{center}
\begin{tabular}{>{\bfseries}lccccccc}
\toprule
\textbf{Method} & \thyroid & \bc & \speech & \pg & \shuttle & \kdd & \textbf{Mean}   \\
\midrule
DIF~\cite{ouardini2019towards} & \underline{83.8}\tiny{$\pm$3.6} & \underline{86.6}\tiny{$\pm$2.1} & 43.9\tiny{$\pm$5.3} & 93.1\tiny{$\pm$0.7} & \underline{98.4}\tiny{$\pm$0.4} & \underline{99.1}\tiny{$\pm$0.1} & 84.1\tiny{$\pm$1.1}\\
MahaAD~\cite{rippel2021modeling} & 74.9 & \textbf{100} & 44.6 & 96.3 & 81.4 & \textbf{100} & 82.9\\
DN2~\cite{bergman2020deep} & 71.9 & \textbf{100} & \textbf{76.7} & \textbf{99.9} & \textbf{99.9} & \underline{99.7} & \underline{91.4} \\
\rowcolor{mygray} NL-Invs & \textbf{96.2}\tiny{$\pm$0.4} & \textbf{100}\tiny{$\pm$0.0} & \underline{71.5}\tiny{$\pm$2.5} & \underline{98.5}\tiny{$\pm$0.1} & 94.9\tiny{$\pm$2.4} & \textbf{100}\tiny{$\pm$0.0} & \textbf{93.5}\tiny{$\pm$0.4}\\
\bottomrule
\end{tabular} 
\end{center}
\end{table*}

Some recent works claim that models pre-trained on ImageNet are not a good foundation for U-OOD detectors because they lead to catastrophic failures on seemingly extremely simple cases (\eg,~CIFAR10:SVHN of task \shiftlowres~\cite{hendrycks2019scaling,yousef2023no}), and argue that U-OOD models should be trained from scratch instead. While we also observe catastrophic failure for \DNtwo~and \CFlow, we find that \NL~is able to reach high AUC without any modification to the underlying neural network. In addition, \MahaAD, \MSCL~and to a lesser extent \DIF{} still reach high scores on \shiftlowres. The presumed failure of models based on pre-trained features for certain tasks might thus be related to other factors, such as incorrect processing of features or inappropriate hyperparameters, rather than an intrinsic inability.

\textbf{\shd.}
Tab.~\ref{tab:shallow} summarizes our results. Here, \MahaAD~is the worst performing method, matching \NL's perfect score on \bc~and \kdd~but struggling on the other datasets. \DIF~achieves good performance except on \speech, although it does not reach a perfect score on any dataset. \DNtwo~performs very well, but \NL~is again the best method overall. 

Overall, there is a clear benefit of \NL~over \MahaAD~on tabular datasets: our non-linear invariants approach improves upon the affine invariants approach by 10.6~percentage points of AUC on average across the six experiments. This large difference compared to the results on \guood{} in Tab.~\ref{tab:best101} suggests that invariants in the deep features extracted from a neural network are linear to some extent. 

\begin{table}[t]
\caption{\textbf{Ablating \NL~on \guood.} Learning non-linear invariants, our backward loss, and $S_{\text{final}}$ are all important for high performance.}
\label{tab:ablation}
\begin{center}
\begin{tabular}{cccc}
\toprule
 \multirow{2}{*}{Invariants} & Scoring & Backwards & \multirow{2}{*}{AUC} \\
 & function & loss &  \\
\midrule
 - & $S_{\text{2NN}}$ & - &  86.2 \\
 Linear & $S_{\text{inv}}$ & - & 86.5 \\
 Non-linear & $S_{\text{inv}}$ & \xmark & 86.9 \\
 Non-linear & $S_{\text{inv}}$ & \cmark & 87.2 \\
 Non-linear & $S_{\text{final}}$ & \cmark & \textbf{87.9}\\
 \bottomrule
\end{tabular}
\end{center}
\end{table}

\subsection{Ablation study}

We ablate our design choices in Tab.~\ref{tab:ablation}. The previous best method, \MahaAD, uses linear invariants and reaches a score of 86.5 AUC on the \guood~benchmark. We find that our generalization of this formulation, which allows for learning non-linear invariants, reaches a new state-of-the-art of 87.2 AUC. Part of this improvement is by means of the backward loss. Furthermore, incorporating $S_{\text{2NN}}$ raises the performance even further to 87.9 AUC.

\subsection{Other architectures}

To show the applicability of \NL~to other architectures and model sizes, we show results on \uniclass~with varying models, including ConvNeXT and a vision transformer, in Tab.~\ref{tab:archs}. All models use the same hyperparameters, and we extract the features from $L=3$ feature maps at the last blocks for all models. In general, models with better performance on ImageNet lead to better U-OOD performance, with ConvNeXT reaching the best results.

\begin{table}[t]
\caption{\textbf{Results for \NL~with different architectures on \uniclass.} All models are pre-trained on ImageNet, with the top-1 column showing the ImageNet top-1 accuracy. \NL~is successful across various architectures and benefits from models with higher top-1 scores.}
\label{tab:archs}
\begin{center}
\begin{tabular}{cccc}
\toprule
 Backbone & Size (M) & top-1 & AUC \\
\midrule
  ResNet18~\cite{he2016deep} & 11.2 & 69.8 & 87.8\\
  EfficientNet-b0~\cite{tan2019efficientnet} & 5.3 & 76.3 & 93.3\\
  ResNet101~\cite{he2016deep} & 42.5 & 77.4 & 93.3 \\
  ViT-B-16~\cite{dosovitskiy2020image} & 86.6 & 81.1 & 93.3\\
  ConvNeXT-B~\cite{liu2022convnet} & 88.6 & 84.1 & 94.4\\
 \bottomrule
\end{tabular}
\end{center}
\end{table}

\subsection{Hyperparameter sensitivity}

\NL~has one main hyperparameter, $p$. We show in Tab.~\ref{tab:sens} that \NL~is robust to the choice of $p$, with its performance changing by as little as 0.3~AUC on \guood~across a wide range of values.

\begin{table}[t]
\caption{\textbf{Hyperparameter sensitivity of \NL.} We report the AUC with a ResNet18 backbone on the \guood~benchmark with different values for its main hyperparameter $p$. \NL~is insensitive to the choice of $p$.}
\label{tab:sens}
\begin{center}
\begin{tabular}{ccccccc}
\toprule
  & 0.5 & 1 & 2 & 5 & 10\\
\midrule
  AUC & 86.6 & 86.7 & 86.8 & 86.6 & 86.5\\
 \bottomrule
\end{tabular}
\end{center}
\end{table}

\subsection{Assessing invariants}

We conduct an additional experiment on CIFAR10 following~\cite{doorenbos2022data} to assess how \NL{} incorporates the intuitive idea of invariants in practice. To this end, we compare how U-OOD methods handle different types of OOD datasets as the number of classes in the training set increases. 

\begin{figure}[t]
  \centering
  \setlength\tabcolsep{5pt}
  \begin{tabular}{cc}
    \includegraphics[width=0.4\linewidth]{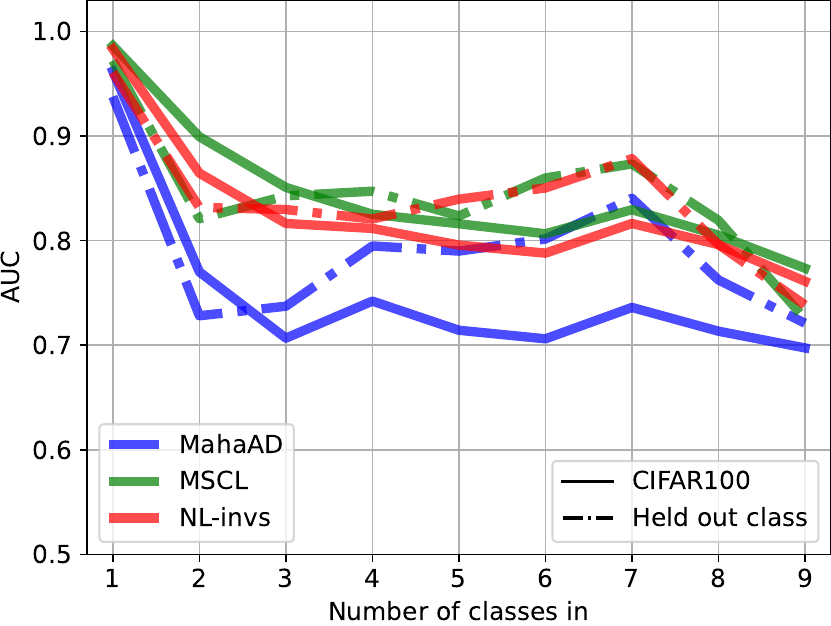} &
    \includegraphics[width=0.4\linewidth]{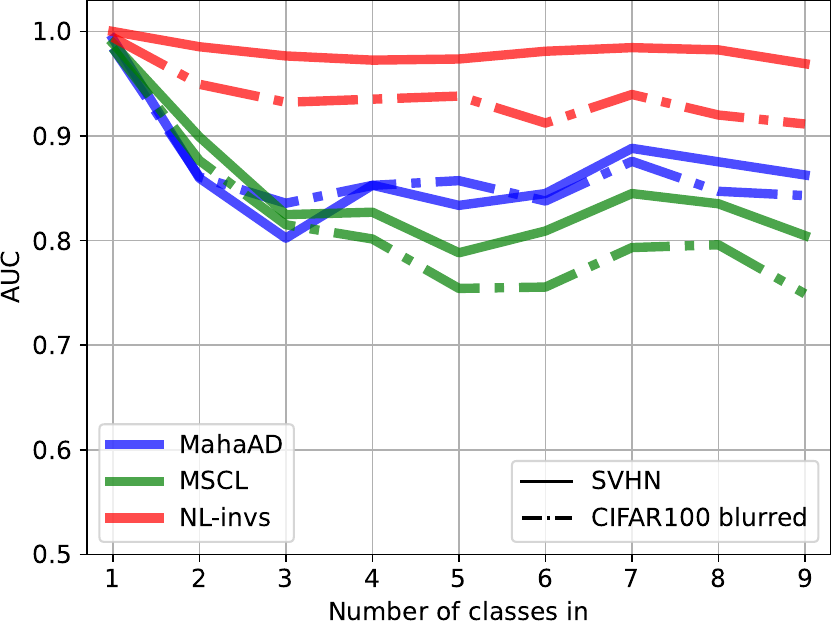} \\
    (a) & (b) \\
  \end{tabular}
    \caption{\textbf{Assessing invariants}. We show how the performance of the top-performing methods changes with respect to the number of classes in the training set for (a)~OOD samples belonging to classes not present in the training data and (b)~visually dissimilar OOD samples. For invariant-based approaches, the AUC remains high when the OOD test set breaks invariants, regardless of the number of classes in the training set.
    }
    \label{fig:vary_classes}
\end{figure}

When the training dataset contains only one class, samples belonging to different classes should be considered outliers, as the class is an invariant. As the number of classes in the training set increases, samples belonging to classes not present in the training dataset should no longer be considered outliers, as the class identity is no longer an invariant. This behavior is shown in Fig.~\ref{fig:vary_classes}(left), where all methods perform as anticipated. Conversely, test samples that exhibit visual dissimilarity from the training set should always be considered outliers, irrespective of the number of classes in the training set. As depicted in Fig.~\ref{fig:vary_classes}(right), our experimental findings indicate that invariant-based methods, namely \MahaAD{} and especially \NL{}, exhibit the expected behavior when test samples come from a different domain, where most of the test samples remain outliers despite the increase in training set classes. In contrast, the next-best performing method, \MSCL{}, experiences a stronger decrease in performance.

\subsection{Loss landscape analysis}

The true U-OOD objective function is impossible to optimize due to the intractability of sampling the entire OOD space. Therefore, all U-OOD methods optimize a proxy loss function to approximate this underlying objective. This, in turn, leads to many U-OOD methods having no apparent correlation between training loss and OOD performance~\cite{reiss2021mean}.

\begin{figure}[t]
    \centering
    \includegraphics[width=0.6\linewidth]{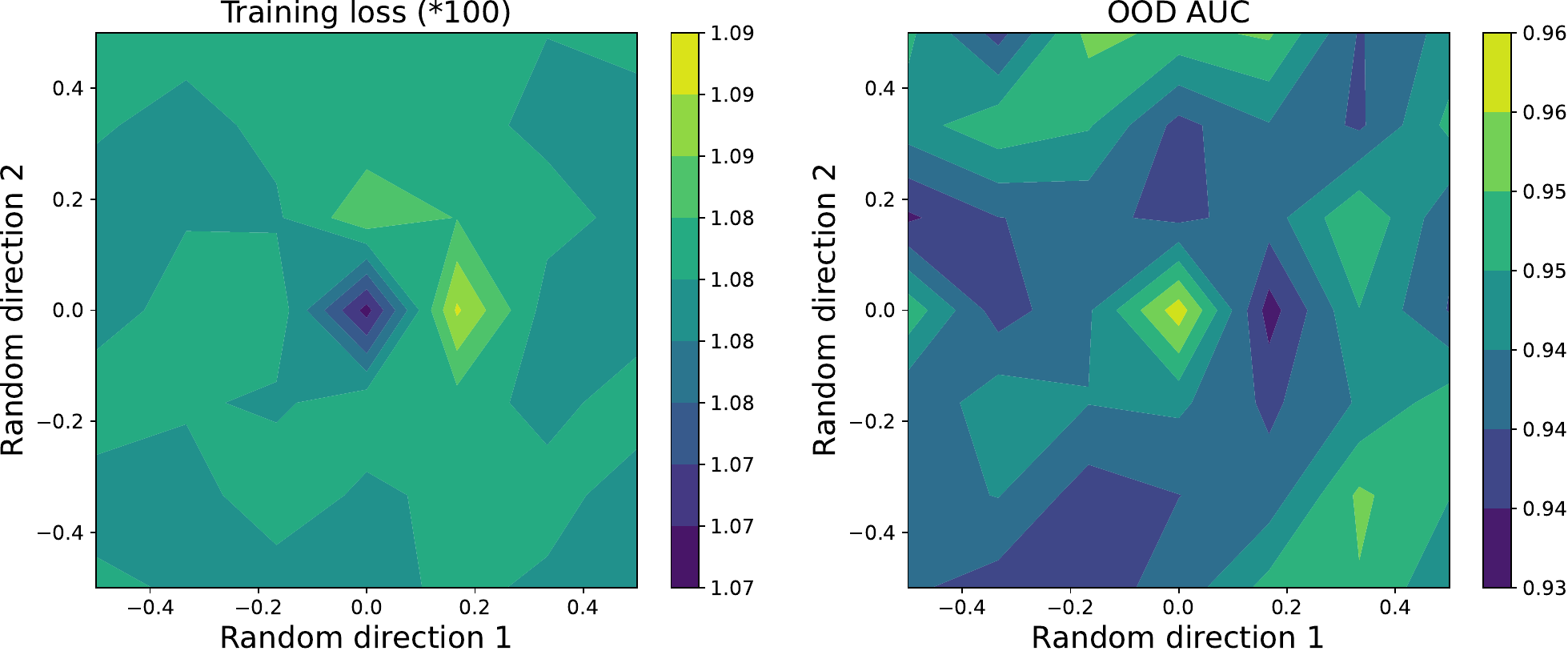}
    \caption{\textbf{Visualizing the loss and AUC landscapes of the VPN.} For \NL, a low training loss corresponds to high U-OOD performance and vice versa.}
    \label{fig:loss}
\end{figure}

Data invariants offer a theoretically sound concept of U-OOD, whereby low training loss regions should correspond to high U-OOD performance and vice versa. To verify this empirically, we utilized \cite{li2018visualizing}'s methodology to visualize training loss and U-OOD AUC along two arbitrary directions in the weight space of the VPN. Our results, displayed in Fig.~\ref{fig:loss} for \texttt{car:rest}, confirm this proposition.

\section{Conclusion}

This work introduces a new U-OOD method that learns data invariants within a training set. Our framework, called \NL, is the first volume-preserving approach to OOD detection. \NL~learns non-linear invariants over a set of training features and generalizes previous invariant-based formulations of U-OOD, reaching state-of-the-art performance when compared against competitive methods on a large-scale benchmark. Additionally, we validate our model on different tabular datasets, showing its generalizability and advantage over affine invariants. 

Finally, we confirm the results of~\cite{doorenbos2022data} and observe that the performance of several U-OOD methods is highly sensitive, with the majority of techniques displaying inconsistent scores across various tasks. Nevertheless, invariant-based approaches maintain a prominent position in terms of consistency, with \NL~outperforming all other methods by achieving the highest overall performance and ranking as the top-performing technique on three out of five tasks on the \guood~benchmark, in addition to obtaining the best score on tabular data. All in all, U-OOD remains challenging due to its many inconsistencies. We believe that with proper evaluation set-ups and theoretically motivated approaches, such as those based on data invariants, significant progress can be made toward the reliable use of deep learning models in everyday settings.

%
%
\bibliographystyle{splncs04}
\bibliography{main}

\appendix
\clearpage
\section{Supplementary Material}

\subsection{Benchmark details}

We provide further details for our \shd~benchmark, which consists of the following six datasets from~\cite{goldstein2016comparative}:

\begin{description}
    \item[\thyroid.] Training:~6'666~samples consisting of 21~measurements of healthy thyroids. Testing:~250~samples from healthy thyroids as inliers and 250 outliers from hyper-functioning and subnormal-functioning thyroids as outliers.
    \item[\bc.] Training:~357~samples with 30 measurements taken from medical images from healthy patients. Testing:~10 inliers from healthy patients and~10 outliers from cancer instances.
    \item[\speech.] Training:~3'625~samples of 400-dimensional features extracted from recordings of people speaking with an American accent. Testing:~61 American-accent recordings as inliers and~61 outliers from speakers with non-American accents.
    \item[\pg.] Training:~719~samples of the digit~`8' represented as a vector of 16 dimensions. Testing:~90~samples of~`8' as inliers and~90~samples of other digits as outliers.
    \item[\shuttle.] Training: 45'586~samples describing a space shuttle's radiator positions with 9-dimensional vectors. Testing: 878~inliers from normal situations, 878~outliers taken from abnormal situations.
    \item[\kdd.] Dataset of simulated traffic in a computer network, where attacks are seen as anomalies and normal traffic as inliers. Training:~619'046~samples of 38~dimensions. Testing:~1'052~inliers from normal traffic and 1'052~outliers.
\end{description}

\subsection{Additional results}

\begin{table*}
\begin{center}
\begin{tabular}{>{\bfseries}lcccccc}
\toprule
\textbf{Method}     & \textbf{\uniclass}& \textbf{\uniano}      & \textbf{\unimed}        & \textbf{\shiftlowres}   & \textbf{\shifthighres}         &\textbf{Mean}   \\
\midrule
CFlow~\cite{gudovskiy2022cflow} & 68.2\tiny{$\pm$0.7} & \textbf{92.9}\tiny{$\pm$0.3} & 65.8\tiny{$\pm$0.3} & 4.3\tiny{$\pm$0.8} & 56.1\tiny{$\pm$0.7} & 57.5\\
DN2~\cite{bergman2020deep} & 79.5 & 82.5 & 73.5 & 19.7 & 66.4 & 64.3\\
DDV~\cite{marquez2019image} & 63.8\tiny{$\pm$1.4} & 62.9\tiny{$\pm$2.8} & 67.2\tiny{$\pm$1.8} & 58.8\tiny{$\pm$3.2} & \textbf{85.8}\tiny{$\pm$2.1} & 67.7 \\
DIF~\cite{ouardini2019towards} & 81.1\tiny{$\pm$0.2} & 78.0\tiny{$\pm$0.5} & 71.8\tiny{$\pm$0.3} & 88.7\tiny{$\pm$2.8} & 75.7\tiny{$\pm$0.9} & 79.1 \\
MahaAD~\cite{rippel2021modeling} & 85.9 & 88.6 & 75.0 & 85.4 & 76.9 & 82.4\\
MSCL~\cite{reiss2021mean} & \textbf{92.7}\tiny{$\pm$0.0} & 86.2\tiny{$\pm$0.1} & \underline{76.5}\tiny{$\pm$0.0} & 90.0\tiny{$\pm$0.2} & 79.9\tiny{$\pm$0.0} & \underline{85.1}\\
\rowcolor{mygray} NL-Invs & \underline{88.5}\tiny{$\pm$0.0} & \underline{89.9}\tiny{$\pm$0.1} & \textbf{77.0\tiny{$\pm$0.1}} & \textbf{97.5}\tiny{$\pm$0.4} & \underline{81.3}\tiny{$\pm$0.3} & \textbf{86.8}\\
\bottomrule
\end{tabular}
\end{center}
\caption{\textbf{Comparative evaluation on \guood.} We report the mean and standard deviation of the AUC over three runs with a ResNet18. Methods without a reported standard deviation are deterministic. \textbf{Bold} and \underline{underlined} indicate best and second best per column, respectively. On aggregate across the experiments, \NL~obtains the best performance.}
\label{tab:best18}
\end{table*}

In~\cite{doorenbos2022data}, all methods were used with their default hyperparameters as provided by their official implementations. However, it is typically unclear how these were selected. Here, we provide results with a unified approach to selecting the hyperparameters of the compared methods.

We tuned all baselines following a consistent protocol where hyperparameters were set via a grid search. We measure the performance of each method on the experiment \texttt{Real A:Quickdraw A} from the task \shifthighres, and select the highest-performing configuration of the grid. 

We briefly describe all baselines and their hyperparameter ranges considered in the grid search. The ranges are based on the values found in each method's original publication. All methods use a ResNet18 pre-trained on ImageNet with images resized to $224\times{}224$. We use the official code for \CFlow\footnote{\url{https://github.com/raghavian/cFlow}}~and \MSCL\footnote{\url{https://github.com/talreiss/Mean-Shifted-Anomaly-Detection}}, and our own implementation for \DNtwo, \DDV, \DIF, \MahaAD~and \NL.
\begin{description}
    \item[CFlow~\cite{gudovskiy2022cflow}] trains multiple conditional NFs on the features of the network, each at a different scale. The condition is given by the spatial location of the features. The final scores are found by aggregating the results at all scales. We search over the number of coupling layers $c\in[4,6,8]$, number of pooling layers $p\in[2,3]$, learning rate  $lr\in[0.0002, 0.00006]$, and batch size $bs\in[32,64,128]$. The best-performing configuration was $c=6, p=3, lr=0.00006, bs=128$.
    \item[DN2~\cite{bergman2020deep}] scores samples with the average distance to the $n$-nearest neighbors in the feature space of the penultimate layer of the network. We search over $k\in[1,2,3,5,10,15,20,25,30]$, where $k=30$ gave the best result.
    \item[DDV~\cite{marquez2019image}] exchanges the final fully-connected layer of the ResNet for a randomly initialized, low-dimensional fully-connected layer. Then, it maximizes the log-likelihood of the training data in this low-dimensional space and scores test samples by their negative log-likelihood. We search over the final-layer dimensionality $d\in[8,16,32,64]$, the Gaussian kernel bandwidth parameter $h\in[-4,-3,-2,-1]$ and the batch size $bs\in[32,64,128]$. The best-performing configuration was $d=8,h=-3,bs=128$.
    \item[DIF~\cite{ouardini2019towards}] concatenates the features extracted from multiple layers of the network and fits an IF to them, which is also used to score test samples. We search over the number of trees $nt\in[100,200,300,400,500]$, the fraction of features used in every tree $mf\in[0.25,0.5,1]$, and the fraction of samples used in every tree $ms\in[0.25,0.5,1]$. The best-performing configuration was $nt=200,mf=0.5,ms=1$.
    \item[MahaAD~\cite{rippel2021modeling}] scores samples by the sum of the Mahalanobis distances computed at multiple locations in the network. It has no hyperparameters.
    \item[MSCL~\cite{reiss2021mean}] combines two losses to fine-tune the final two blocks of the network: the first is an adapted version of the contrastive loss, and the second is the center loss on the normalized features. Test samples are scored using $k$-NN on the normalized features, thus using the cosine similarity as the distance metric. We search over the temperature $t\in[0.05,0.1,0.2,0.25,0.3,0.4]$, the batch size $bs\in[32,64,128]$, and learning rate $lr\in[10^{-4},5\cdot10^{-5},10^{-5}]$. The best-performing configuration was $t=0.05,bs=128,lr=5\cdot10^{-5}$.
    \item[NL-Invs] learns non-linear invariants over the training features at multiple scales and uses those to score test samples. We search over $p\in[1, 2, 5, 10]$, which is the hyperparameter defining the largest number of principal components of the data that jointly explain less than $p$\%~of the variance, which is used per layer to set the $K_\ell$'s. We also search over batch size $bs\in[32,64,128]$, and learning rate $lr\in[0.01, 0.005, 0.001]$. The best-performing configuration was $p=2,bs=32,lr=0.01$.
    
\end{description}

From Tab.~\ref{tab:best18}, we see that also with these settings, \NL~reaches the state-of-the-art performance on the \guood~benchmark, scoring best on two of the five tasks and second best on the remaining three.

\end{document}